# Deep learning on rail profiles matching

Kunqi Wang[1], Daolin Si[1], Pu Wang[1], Jing Ge[1], Peiyuan Ni[2*], Shuguo Wang[1*]

1) Railway Engineering Research Institute, China Academy of Railway Sciences Corporation Limited, Beijing, China

2) Advanced Robotics Center, National University of Singapore, Singapore

*Corresponding author:

Peiyuan Ni (pyni@nus.edu.sg)

Shuguo Wang (zzddxx4473@sina.com)

**Abstract**

Matching the rail cross-section profiles measured on site with the designed profile is a must to evaluate the wear of the rail, which is very important for track maintenance and rail safety. So far, the measured rail profiles to be matched usually have four features, that is, large amount of data, diverse section shapes, hardware made errors, and human experience needs to be introduced to solve the complex situation on site during matching process. However, traditional matching methods based on feature points or feature lines could no longer meet the requirements. To this end, we first establish the rail profiles matching dataset composed of 46386 pairs of professional manual matched data, then propose a general high-precision method for rail profiles matching using pre-trained convolutional neural network (CNN). This new method based on deep learning is promising to be the dominant approach for this issue. Source code is at https://github.com/Kunqi1994/Deep-learning-on-rail-profile-matching.

Keywords: Deep learning, Pretrained convolutional neural networks, Rail wear, Rail profiles matching

**Introduction**

The development of high-speed rail plays an important role in China's domestic economy and modernization drive. Different from other transportation, wheel-rail interaction is the heart of railway and the core factor determining the stability and safety of train [1]. After long-term operation, the rail begins to wear gradually and the geometric profile of the rail changes. In particular, the rail parts in the turnout area of the track, due to the irregularity of track geometry and the impact of train, are most vulnerable to wear [2, 3]. When the vertical wear and slide wear of rail exceed the specified safety value, the wheel-rail relationship deteriorates. At this time, if the rail profile is not ground or the rail components are not replaced in time, not only the stability but also the safety of the train will be endangered. Among this, match the measured rail profile with the designed profile and then calculate the rail wear is a key and essential step.

Speaking of rail profile matching, first, we need to understand the characteristics of the profile matching issue. To the best of knowledge, the rail profile to be matched usually has four features. First, the amount of the measured profiles to be matched is extremely large, at least ten million in China every year. With the promotion of intelligent maintenance and repair in the railway, the amount of data will continue to increase in the future. Second, there are many shapes of rail section profiles, including not only the typical rail profiles, but also the rail profiles with variable area sections such as switch, frogs and combined profile in the turnout area [4], as shown in Figure 1a. Third, because of the limitation of rail profile measurement hardware, sometimes there are hardware-made errors in the measured profile, and sometimes only the section information above the rail waist is measured, as shown in Figure 1b. Last, professional manual experience needs to be introduced to deal with the complex situation of the railway site during matching process. For example, as shown in Figure 1c, in many cases, the working edge and non-working edge of the rail section profile do not coincide.

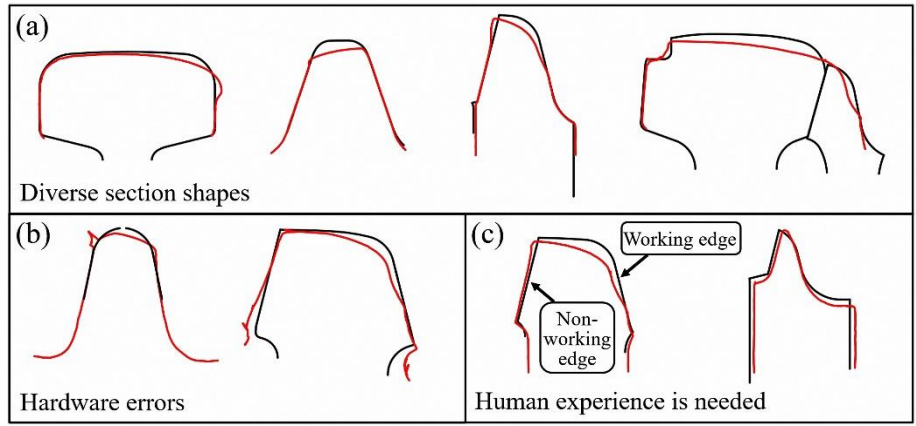

Figure 1. Features of rail cross-section profiles matching. (a) diverse section shapes (b) hardware errors (c) manual experience needs to be introduced. The black and the red line are the designed and the measured profiles respectively. The working edge and non-working edge of the profile are also illustrated here.

At present, there are two solutions for rail profile matching. One is that match through professionals, the matching effect is very good, but the matching efficiency is very low. Considering that the amount of profile data to be matched is too large, and for another, different people have different profile matching criteria, thus it is difficult to form standards, it is obvious that this method can no longer meet the requirements. The second method is using traditional matching method based on feature points and feature lines [5-7]. For example, Wang et al. proposed a method for automatic extraction of small circular arc of the rail waist and achieved the high-precision matching of rail profile [5]. Feng et al. achieved accurate matching of two-dimensional point cloud of rail profile by improving the ICP (Iterative Close Point) algorithm [6]. Furthermore, in addition to ICP algorithm, image matching algorithm based on image processing, including RANSAC algorithm [8] and least square algorithm [9, 10] and other methods [11, 12], might also be used to solve the rail profile matching issue. Though the advantage of the second method is of high matching accuracy and efficiency, but they could not introduce manual experience to solve the complex situation on the site during matching process.

Aiming at the issue of rail profile matching, this paper first establishes the rail profile matching dataset, that is, 46386 pairs of professional manual matched data are collected. Then, we propose a model to solve the profile matching using pre-trained convolutional neural network method in deep learning. The advantages of this new rail profile matching method are: (1) Professional manual profile matching experience can be introduced to solve the situation that the working edge and non-working edge of the profile do not coincide, or on hardware made errors. (2) It is not necessary to know all the section information of the measured profile, even the rail waist. (3) It can match various shapes of section profiles, including typical section profile, switch profile and frog profile and so on.

## 2. Dataset

Before preparing the dataset, it is a must to understand the process of calculating rail wear manually, normally including four steps. First, find the designed profile or initial profile of the rail section, and the profile data is usually in dwg format. Second, import the same rail section profile measured by hardware on the railway site into the same dwg file. Third, a core step in the whole profile matching process is, translate the measured profile in X and Y directions in the dwg file through professional manual experience to match the designed profile. The schematic diagram of the matching process is shown in Figure 2. Last, for the matched rail profile, according to the definition of rail wear, the vertical wear and side wear of rail are calculated, and the grinding and replacement time of rail parts are determined.

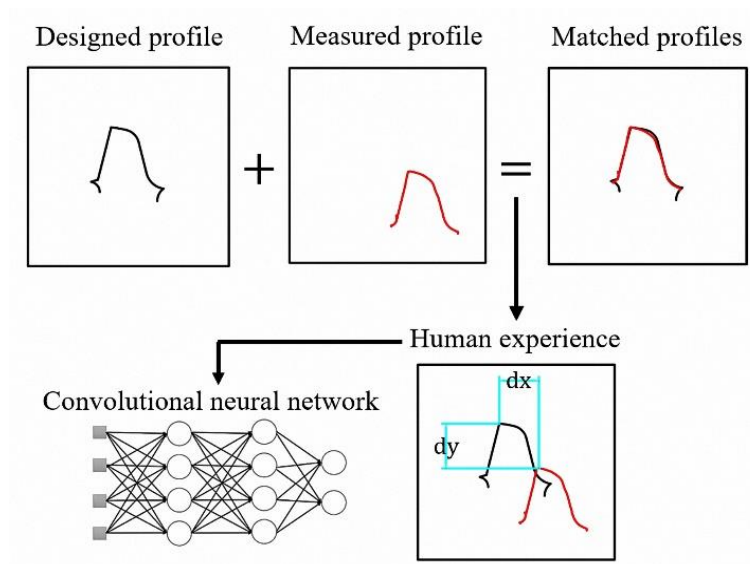

Figure 2. Schematic diagram of solving the problem of rail section profiles matching manually and the alternative solution using convolution neural network in deep learning.

Matching through professionals has many advantages, but when the amount of data is large, the efficiency will be very low, especially the profile data of rail section to be matched in China is close to 10 million every year. Therefore, can we use deep learning to perform the above core matching process? When using deep learning to deal with profile matching, how to make neural network better learn professional manual matching experience is an important issue. Considering that image can well capture the spatial features of the profiles, we choose to use the convolution neural network method to solve this problem, as shown in Figure 2.

The construction of the dataset includes images containing the information of the designed profile and the measured profile, and its corresponding labels. For the image input to the convolutional neural network, its size is 512 * 512. The background is white, the designed profile is black, and the measured profile is red, as shown in Figure 3. Each image corresponds to the actual size of 153.6mm, and one pixel corresponds to 0.3mm resolution. In each image, the centroids of the designed profile and the measured profile are randomly distributed in a centered square with a side length of 40mm. Since when the profile centroid distribution is greater than 40mm, some profile data may exceed the boundary of the whole image. It should be emphasized here that this method does not need to know all the information of rail section profile, or even the rail waist, as shown in the red profile in Figure 3. In addition, the displacement of the measured profile relative to the initial profile in the X and Y directions is rescaled to - 1 to 1 of the label.

The dataset consists of 46386 samples. These samples were divided into training set (~70 %), validation set (~15 %) and test set (~15 %). The specific numbers are 32480, 6976 and 6930 respectively. The training set was used to learn the model parameters, the validation set was necessary for the hyperparameter optimization and with the test set the performance of the model is evaluated on unseen data. When splitting the data, it must be guaranteed that different profile shapes, such as typical section profiles, switch section profiles and frog section profiles, are randomly distributed at every set.

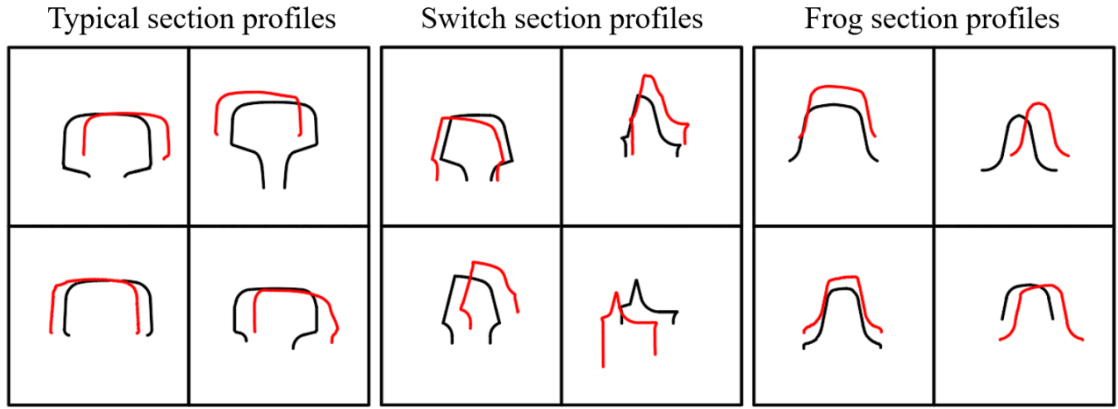

Figure 3. The sample pairs randomly selected in the dataset include typical section profiles, switch section profiles and frog section profiles. The black and the red line are the designed and the measured profiles respectively.

3. Methods and results

Here, we first need to determine the input, output and architecture of the model. As mentioned in the Dataset section, the input of the model is an image (512*512 pixel) including both the designed profile and the measured profile. Each of the image was rescaled to 224 x 224 pixels. The output of the model is two values, that is, the relative translation (dx1, dy1) in the X and Y directions. The architecture of the model used is shown in Figure 4a. The core of the model is a single channel pre-trained Resnet18 network with the classification category being 2 [13]. The used hyperparameter of the model during training process are shown in Figure 4b. The weights of the network were initialized with the pre-trained parameters [13] in ImageNet dataset and updated with the Adam algorithm [14]. The learning rate was set to 0.0001. The batch size was 32. The epoch number was 1000 and large enough for the present issue. In addition, the output layer of the model does not use the activation function, and any layer of the pre-trained Resnet18 model parameters is not frozen.

The mean square error (MSE) was used as loss function during training process and to evaluate the quality of the regression on the validation set. Considering the high precision requirement of profile matching, it is very important to formulate the criterion of successful matching. Here, we define the professional manual matching data in the dataset as the correct result, and the corresponding label values are dx and dy. If the model outputs dx1 and dy1 meet the relationship |dx1-dx|<0.4mm and |dy1-dy|<0.4mm, it indicates a set of data is successfully matched. It should be noted here that 0.4mm is a reasonable accuracy for rail profile matching. The model was implemented in Python using the libraries Pytorch [15]. It was trained on the GPU RTX A4000 graphical processing unit (NVIDIA corporation). The training time for one case was normally about seven days.

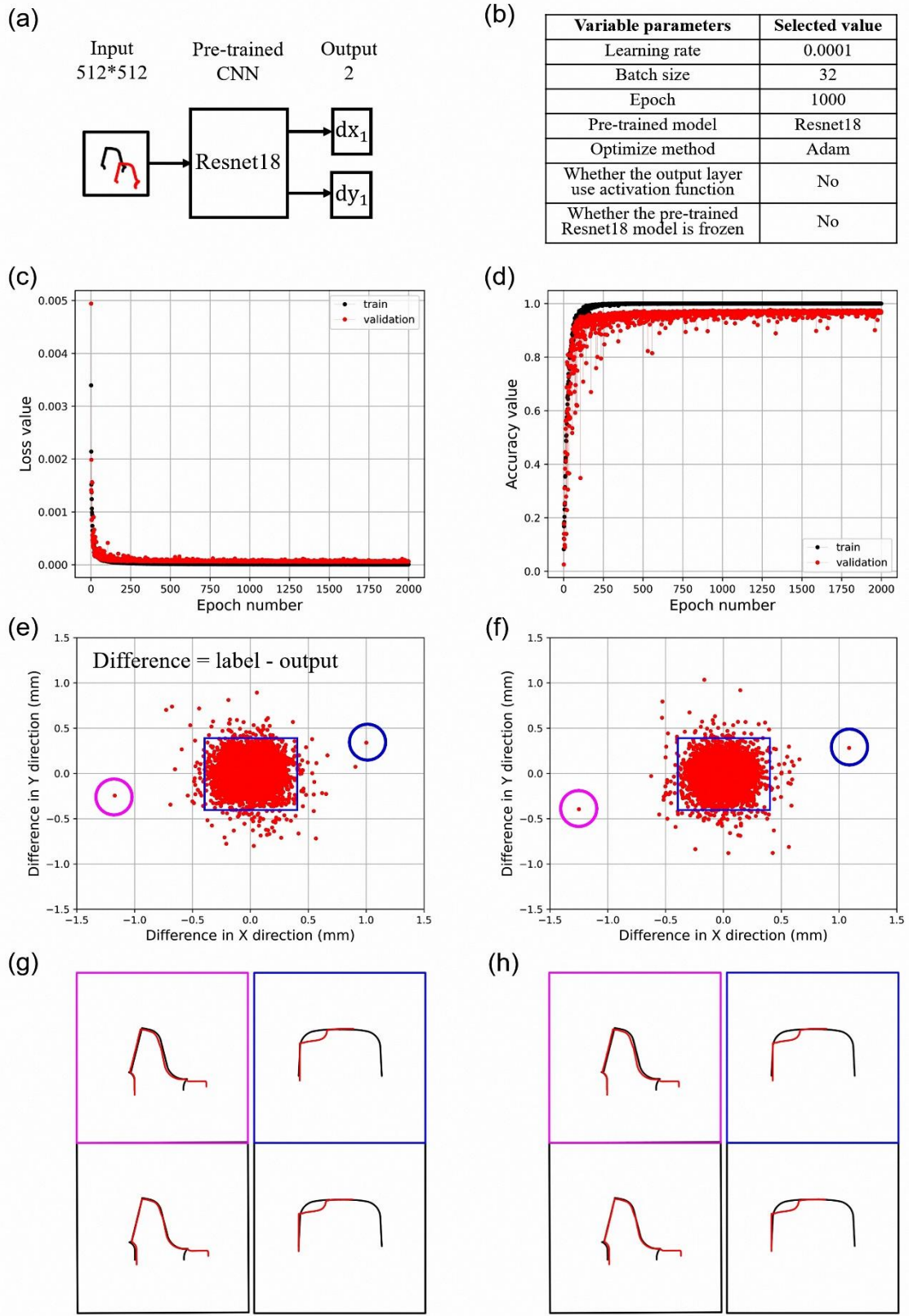

**Figure 4. Results of the pre-trained single channel Resnet18 model.** (a) Schematic diagram of the single channel model architecture. (b) The chosen hyperparameters of the model during the training process. (c) Loss versus epoch in the training and validation process. (d) Matching accuracy versus epoch in the training and validation

process. (e) Prediction results of using the pre-trained Resnet18 model in the test dataset. (f) Average four different prediction results using pre-trained Resnet18 model in the test dataset. (g-h) Two specific prediction results corresponding to blue and pink circles in Figure 4(e-f) are illustrated here. The ideal matching result is circled with a black border below.

The prediction results of the model during training process are shown in Figure 4c. With the increase of epoch number, the loss value of both the train and the validation first decreases sharply and then tends to zero. During this process, the validation loss is still slightly greater than the train loss. Meantime, as shown in Figure 4d, with the increase of epoch number, the accuracy value (success prediction matching rate) on the train set rises sharply and then converge to 100%, but on the validation set, although the accuracy also rises sharply, the highest prediction accuracy in validation dataset is 0.9791. This value can never reach 100%, even if the epoch number increases to 5000, and the possible reason is that the amount of data is not large enough to support the model to have a strong generalization ability.

The predicted results on test set are shown in Figure 4e, the red point shows the difference between the prediction result and the label in X and Y directions. According to statistics, the accuracy value of successful matching is 0.979 (in the blue frame). Obviously, the current model could not solve the rail profile matching issue. To better analyze of the prediction results, we select the two results with the worst matching effect (pink and blue circles in Figure 4e) for display. The corresponding results are shown in Figure 4g. The results predicted by the model are in the pink and blue boxes, and the corresponding professional manual matching results are in the black box below. It can be seen that the matching effect on these two samples is not satisfied.

In order to improve the matching accuracy, we use the average of **four** different single channel Resnet18 prediction results on test set. Why we choose four to average is because the maximum difference is less than 1.6mm, nearly four times that of 0.4mm. Specially, we use the same model and the same hyperparameters to train the model four times, first find and save the best model parameters in validation set and then to predict on test set every time. The results are shown in Figure 4f, the red point shows the difference between the prediction result and the label in X and Y directions. The accuracy value of successful matching is 0.9737 (in the blue frame). Therefore, even after average, the matching success rate still could not meet the on-site requirements. We also selected the worst case of prediction results to display, as shown in Figure 4h, we found that the samples of worst prediction accuracy are the same as those in Figure 4g.

**4. discussion**

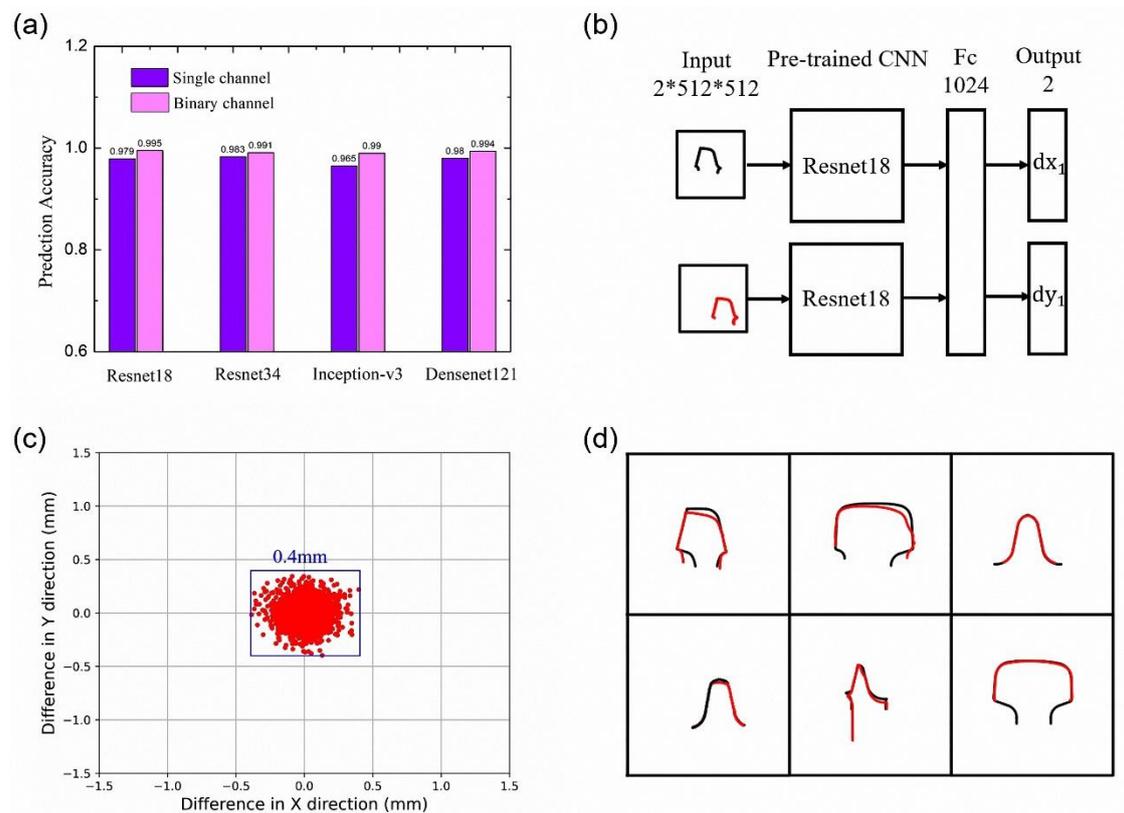

**Figure 5. Exploration on improving matching accuracy.** (a) Comparison of matching accuracy of single and binary channel architecture, both of which include Resnet18, Resnet34, GoogLenet-v3, Densenet121 algorithm. (b) Schematic diagram of the binary channel architecture, Resnet18 model as an example here. (c) Average the prediction results of three models include binary channel Resnet18, single channel Resnet18 and single channel GoogLenet-v3. (d)Display of randomly selected six prediction results corresponding to Figure 5c.

The results in Figure 4 show that the single channel Resnet18 model could not achieve 100% matching accuracy, here two other approaches are adopted to improve the prediction accuracy. The first is to change the core algorithm after the single channel network structure model is determined, such as Resnet34 [13], GoogLenet-v3 [16] and Densenet121 [17], in which all the hyperparameters remain the same as those in Figure 4b during the training process. The second is, inspired by the binary camera matching [18, 19], we choose to use the binary network structure to improve the matching accuracy, and the architecture of the model is shown in Figure 5b, binary channel Resnet18 is as an example. Different from the single channel Resnet18 algorithm in Figure 4a, its input of binary channel Resnet18 model is not one picture, but two pictures contain the designed profile and the measured profile respectively, and the corresponding dataset changes accordingly. Besides, a full connection layer is added to merge the prediction results of the two single channel. It should be noted that the training hyperparameters are the same as those in Figure 4b, and the initial parameters of the pre-trained model are also directly used here.

The prediction results are shown in Figure 5a. It can be seen that the prediction accuracy for the single channel corresponding to Resnet18, Resnet34, GoogLenet-v3 and Densenet121 are 0.979, 0.983, 0.965 and 0.98 respectively, and none of them reaches 100%. Interestingly, for the corresponding binary channel model, the prediction accuracy is 0.995, 0.991, 0.99 and 0.994 respectively. These results show that, on the one hand, the matching accuracy of binary channel Resnet18 is the highest among these, reaching 0.995. On the other hand, the prediction results of all binary channel models are better than the relative single channel. So far, we have found that for one model, whether single or binary channel, the accuracy of profile matching could not meet the requirements on site, that is, the matching accuracy needs to reach 100%, since the matching accuracy is a major matter related to train safety. Therefore, here we attempt to improve the matching accuracy by stacking different models.

Here, we choose to use the weighted average method, and the specific weight given by binary channel Resnet18, single channel Resnet18 and googLeNet-v3 are 0.5, 0.25 and 0.25 respectively. The reason for choosing **three** methods to average is that for the binary channel Resnet18 algorithm with the best matching accuracy, the difference between the worst predicted matching result and the label is larger than 0.8mm and smaller than 1.2mm in X or Y direction. Besides, we choose to use weighted average method instead of direct average is that binary channel Resnet18 algorithm is the best and should be given a larger weighted proportion. The predicted result is shown in Figure 5c, the red point shows the difference between the prediction result and the label in X and Y directions, which are all less than 0.04mm (in the blue frame). In other words, the matching accuracy finally reaches 100%. Further, to visually observe the matching results, 6 samples in the test dataset were randomly selected, as shown in Figure 5d. Inside each black box is a sample, in which the black line is the designed profile and the red line is the measured profile. Through manual verification, the matching result of the algorithm is reasonable and of high-precision, consistent with that of manual matching.

## 5. Conclusion and outlook

In this paper, the pre-trained convolutional neural network in deep learning is introduced into the rail cross section profile matching for the first time. Specially, the dataset composed of 46386 pairs of professional manual matched data is constructed, then the architecture of the matching model is proposed, and finally the algorithm on how to improve the profile matching accuracy is discussed. The matching algorithm proposed can not only achieve the accuracy of the traditional matching algorithm based on feature points or feature lines, more importantly, it could introduce the professional manual matching experience to solve the complex situation on site. That is, it can not only solve the situation when both the working edge and non-working edge of profile do not coincide, but also, the measured profile to be matched does not need to know the full section profile or even rail waist profile. In addition, it can match the rail profiles of various complex shapes. This new profile matching method based on deep learning is expected to become the mainstream of rail profile matching and advances the realization of intelligent railway track detection and maintenance. It can be applied to many hardware equipment, such as rail profile measuring instrument, track geometry detection equipment and rail grinding vehicle.

Further interesting steps on improving matching performance include, first, find a better algorithm in the binary network structure model or use Transformer [20,21] to improve the profile matching accuracy. Second, more data should be accumulated and combined rail profiles data should be added for the dataset. Third, network compression [22] is needed to reduce the model parameters and improve the processing speed.

**Acknowledgements**

S.G.W wishes to acknowledge the financial support by the National Natural Science Foundation of China (No.51878661). D.L.S wishes to acknowledge the financial support by the National Key R&D Program of China (No. 2021YFB3703600).

**Reference**

[1] W.M. Zhai, K.Y. Wang, C.B. Cai, Fundamentals of vehicle-track coupled dynamics, Vehicle System Dynamics, 47 (2009) 1349-1376.
[2] H. Xu, P. Wang, D. Liu, J.H. Xu, R. Chen, Maintenance Technologies Research of High speed Turnout, International Conference on Civil Engineering and Transportation (ICCET 2011), Jinan, PEOPLES R CHINA, 2011, pp. 114-120.
[3] J.M. Xu, P. Wang, L. Wang, R. Chen, Effects of profile wear on wheel-rail contact conditions and dynamic interaction of vehicle and turnout, Advances in Mechanical Engineering, 8 (2016).
[4] J. Lai, J.M. Xu, P. Wang, Z. Yan, S.G. Wang, R. Chen, J.L. Sun, Numerical investigation of dynamic derailment behavior of railway vehicle when passing through a turnout, Engineering Failure Analysis, 121 (2021).
[5] H. Wang, S. Wang, W. Wang, Automatic registration method of rail profile in train-running environment, J. Beijing Univ. Aeronaut. Astronaut. (China), 44 (2018) 2273-2282.
[6] K. Feng, L. Yu, D. Zhan, D. Zhang, Research on Fast and Robust Matching Algorithm in Inspection of Full Cross-section Rail Profile, Journal of the China Railway Society, 41 (2019) 162-167.
[7] D. Zhan, L. Yu, J. Xiao, M. Lu, Study on Dynamic Matching Algorithm in Inspection of Full Cross-section of Rail Profile, Journal of the China Railway Society, 37 (2015) 71-77.
[8] M.A. Fischler, R.C. Bolles, RANDOM SAMPLE CONSENSUS - A PARADIGM FOR MODEL-FITTING WITH APPLICATIONS TO IMAGE-ANALYSIS AND AUTOMATED CARTOGRAPHY, Communications of the Acm, 24 (1981) 381-395.
[9] D.G. Lowe, Distinctive image features from scale-invariant keypoints, International Journal of Computer Vision, 60 (2004) 91-110.

[10] D.W. Marquardt, AN ALGORITHM FOR LEAST-SQUARES ESTIMATION OF NONLINEAR PARAMETERS, Journal of the Society for Industrial and Applied Mathematics, 11 (1963) 431-441.
[11] M. Cui, J. Femiani, J. Hu, P. Wonka, A. Razdan, Curve matching for open 2D curves, Pattern Recognition Letters, 30 (2009) 1-10.
[12] S. Elghoul, F. Ghorbel, An Efficient 2D Curve Matching Algorithm under Affine Transformations, 13th International Joint Conference on Computer Vision, Imaging and Computer Graphics Theory and Applications (VISIGRAPP) / International Conference on Computer Vision Theory and Applications (VISAPP), Funchal, PORTUGAL, 2018, pp. 474-480.
[13] K.M. He, X.Y. Zhang, S.Q. Ren, J. Sun, Ieee, Deep Residual Learning for Image Recognition, 2016 IEEE Conference on Computer Vision and Pattern Recognition (CVPR), Seattle, WA, 2016, pp. 770-778.
[14] B.J. Kingma D P, Adam: A method for stochastic optimization, DOI (2014).
[15] A. Paszke, S. Gross, F. Massa, A. Lerer, J. Bradbury, G. Chanan, T. Killeen, Z.M. Lin, N. Gimelshein, L. Antiga, A. Desmaison, A. Kopf, E. Yang, Z. DeVito, M. Raison, A. Tejani, S. Chilamkurthy, B. Steiner, L. Fang, J.J. Bai, S. Chintala, PyTorch: An Imperative Style, High-Performance Deep Learning Library, 33rd Conference on Neural Information Processing Systems (NeurIPS), Vancouver, CANADA, 2019.
[16] C. Szegedy, V. Vanhoucke, S. Ioffe, J. Shlens, Z. Wojna, Ieee, Rethinking the Inception Architecture for Computer Vision, 2016 IEEE Conference on Computer Vision and Pattern Recognition (CVPR), Seattle, WA, 2016, pp. 2818-2826.
[17] G. Huang, Z. Liu, L. van der Maaten, K.Q. Weinberger, Ieee, Densely Connected Convolutional Networks, 30th IEEE/CVF Conference on Computer Vision and Pattern Recognition (CVPR), Honolulu, HI, 2017, pp. 2261-2269.
[18] J. Xiao, H. Tian, W. Zou, L. Tong, J. Lei, Stereo MatchingBased on Convolutional Neural Network, Acta Optica Sinica, 38 (2018) 0815017.
[19] P.Y. Ni, W.G. Zhang, W.B. Bai, M.J. Lin, Q.X. Cao, A New Approach Based on Two-stream CNNs for Novel Objects Grasping in Clutter, Journal of Intelligent & Robotic Systems, 94 (2019) 161-177.
[20] A. Vaswani, N. Shazeer, N. Parmar, J. Uszkoreit, L. Jones, A.N. Gomez, L. Kaiser, I. Polosukhin, Attention Is All You Need, 31st Annual Conference on Neural Information Processing Systems (NIPS), Long Beach, CA, 2017.
[21] Dosovitskiy A, Beyer L, Kolesnikov A, et al. An image is worth 16x16 words: Transformers for image recognition at scale[J]. arXiv preprint arXiv:2010.11929, 2020.
[22] L. Deng, G.Q. Li, S. Han, L.P. Shi, Y. Xie, Model Compression and Hardware Acceleration for Neural Networks: A Comprehensive Survey, Proceedings of the Ieee, 108 (2020) 485-532.